# Enhancing Semantic Word Representations by Embedding Deeper Word Relationships

Anupiya Nugaliyadde, Kok Wai Wong, Ferdous Sohel and Hong Xie
School of Engineering and Information Technology, Murdoch University, Western Australia.
a.nugaliyadde@murdoch.edu.au, k.wong@murdoch.edu.au, f.sohel@murdoch.edu.au,
h.xie@murdoch.edu.au

**ABSTRACT**

Word representations are created using analogy context-based statistics and lexical relations on words. Word representations are inputs for the learning models in Natural Language Understanding (NLU) tasks. However, to understand language, knowing only the context is not sufficient. Reading between the lines is a key component of NLU. Embedding deeper word relationships which are not represented in the context enhances the word representation. This paper presents a word embedding which combines an analogy, context-based statistics using Word2Vec, and deeper word relationships using Conceptnet, to create an expanded word representation. In order to fine-tune the word representation, Self-Organizing Map is used to optimize it. The proposed word representation is compared with semantic word representations using Simlex 999. Furthermore, the use of 3D visual representations has shown to be capable of representing the similarity and association between words. The proposed word representation shows a Spearman correlation score of 0.886 and provided the best results when compared to the current state-of-the-art methods, and exceed the human performance of 0.78.

## 1. INTRODUCTION

Most Natural Language Understanding (NLU) models use word representations for their inputs [3]. Word representations are used for question answering [5] [6], machine translation [7], dialog systems [8], text understanding [9], and named entity recognition [10].

The commonly used word representations are those created using semantic representation and the lexical information of the local context and the global context [3]. A semantic word representations can demonstrate, "King to queen is to man to woman", (king-queen=man-woman) [11]. GloVe [12] and Word2Vec[1] are the more commonly used word embeddings, due to their capability of holding a semantic relationship. GloVe uses a global word co-occurrence count, which uses the statistic of the corpus and the semantics of the words in the corpus to create a vector for the words [12]. Word2Vec, a predictive model, uses the context words to predict the target word. Word2Vec uses a feedforward network [1] for the prediction. Both GloVe and Word2Vec use large text corpus to create the context-derived word representations.

NLU relies on understanding the context by using information not expressed in the context, which normally requires reading between the lines. Communication requires more information than the current context which is believed to be known by the listener [13]. Retaining more information in a learning model is shown to improve performance over models that do not retain information [14]. Conceptnet is a general knowledge graph comprising of words and their relationships [15]. It contains the knowledge that can be provided for better word representations. However, there is a limitation on containing the semantic knowledge, which is also important in word embedding.

This paper presents a model which combines semantic knowledge and knowledge-graph extracted knowledge for a given word to create word representations. The resulting word representation contains more relationships and knowledge than semantic models. A higher weight is given to the semantic word representation to avoid over-generalization of the final word representation. The proposed model is evaluated using SimLex-999 [16]. Furthermore, to illustrate the capability of differentiating association and similarity, the word embedding is represented in 3D graphs as presented in this paper.

The contributions of the paper are to i) provide a method to generate a word representation that combines Semantic knowledge and knowledge-graph information, ii) evaluate and demonstrate the ability of such a word representation in capturing both similarity and association between words.

## 2. BACKGROUND

Word representations have moved from lexical based to semantic-based to provide information-rich word representations. The early word representations, Latent Semantic Analysis (LSA) hold the statistical representations on the corpus. LSA is a low-dimensional word representation based on the term-document frequency. These models fail at co-occurrences with general words (e.g. and, a, the, etc.).

Later, Positive Pointwise Mutual Information (PPMI) [17] and Hellinger Principal Component Analysis (PCA) [18] were used to generate vector representations for words. The word representation was generalized and could be used for many natural language processing tasks. However, these cannot be used for language understanding [19]. The semantic-based vector representations have shown the potential for improving the understanding of learning models [5]. The semantic word representations produce dimensions of meaning and hold the distributed representation for the words. Continuous Bag-of-Words (CBoW) and skip-gram model shows linguistic patterns as linear relationships between the word vectors. This linear relationship is used to demonstrate the related words in a word representation. Learning based natural language processing applications using this representation would not receive the full meaning representation of the words since it does not use the co-occurrence statistics of the corpus.

GloVe addresses the problems faced by the skip-gram model which did not focus on the statistics of the corpus. In addition, GloVe has the capability of achieving analogy tasks. Therefore, based on the context, GloVe supports a higher level of meaning representation [12]. Word2Vec also uses the co-occurrence statistics from the corpus [20] and the semantics of the corpus. Word2Vec and GloVe are contexts depended. Therefore, depending on the context the vector representation has drastic changes. The word representations and the relationships are based only on the textual context. The word representation does not hold information that is not available in the context.

Conceptnets' knowledge graph is used to create word representation [21] and language understanding [14]. Furthermore, to support language understanding, Conceptnet is used for word representations, using PPMI and expanded retrofitting [22]. The PPMI creates the term-term matrix from the corpus text. The context of a term would be the terms that appear nearby in the text by using sparse matrices. The sparse matrices are used to derive word embedding. The expanded retrofitting using Concepnet uses the multilingual connections to learn English words from language translation. However, the word representations do not consider context-based statistics and analogy. Therefore, it loses the information embedded in a textual context.

Semantics is a key element to provide an understanding of the textual context. However, understanding context does not rely only on the textual context [23]. NLU requires the capability of understanding analogy and corpus statistic. However, to gain further knowledge and understanding deeper relationships that are not visible in the context but known by the listener is required. Conceptnet was developed with the intention of providing the general knowledge for natural language understanding tasks [15]. Conceptnet provides related words and relationships connected to a given word. These connected words and relationships provide hidden knowledge on the textual context. However, the Conceptnet does not provide complete information such as corpus statistic to provide complete natural language understanding. Therefore, it is the aim of this paper to propose a better word embedding.

## 3. METHODOLOGY

This section discusses word embedding that takes into account the semantic relationships and the relationships which are not visible in the textual context. GloVe and Word2Vec use word embedding based on semantic relationships. However, the word embedding is context dependent. To avoid context dependencies and to allow a general representation, Conceptnet and Word2Vec are combined. This combination makes it possible of expressing similarity and association for word embedding which is not available in Word2Vec and GloVe.

The proposed word embedding gives the word w with a final vector representation of $v_f$, which encompass more complete relationship of $w$ with a deeper relationship with other words. $v_f$ combines a semantic representation and a context independent representation for a given word. Therefore, this embedding is capable of representing the embedded meaning even though it is not visible in the textual context.

The embedding creates a 300-row vector for each word representation. 300-row vector was chosen as it presented the best results from the experiments that were conducted by comparing 50, 100, 300, and 500 vector sizes. The embedding consists of the following two main embedding; semantic word embedding, and generalized word embedding.

The semantic and knowledge-graph embedding are combined to create one-word embedding, representing both a semantic and general word representation. These two embeddings are necessary to clearly understand words in a given context. Figure 1 presented a summary of the steps that will be described as follows.

### 3.1 Semantic Word Embedding

Semantic word embedding is used to embed the meaning expressed through the textual context. Semantic word embedding is generated through Word2Vec. Equation (1) gives the word ($w$) is embedded using Word2Vec to create an initial word embedding which holds the word similarity. The vector ($v$) holds the word similarity depending on the context the word appears.

$$v = Word2Vec(w) \qquad (1)$$

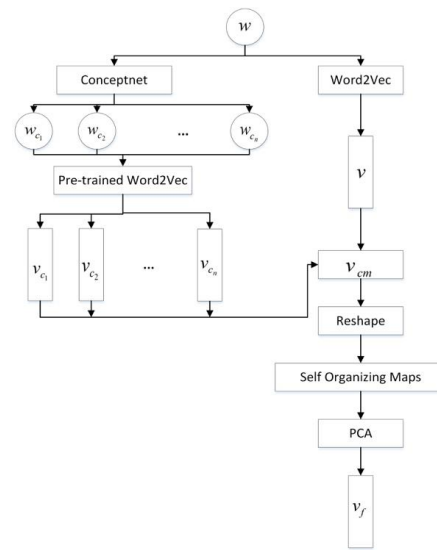

**Figure 1. Overall methodology and data flow for the word embedding. The word $w$ is passed on to Conceptnet and Word2Vec. Word2Vec creates the word embedding $v$. The words extracted from Conceptnet $w_{c_n}$ are passed on to a pre-trained Word2Vec to generate $v_{c_n}$ and combine with $v$ using $\sum_{i=1}^{n} v_{c_i} \cdot v$, to generate $v_{cm}$. $v_{cm}$ is reshaped to a 2D matrix. The 2D matrix is passed to a SOM and the generated matrix is then passed on to PCA for dimensional reduction and transformed to the final vector $v_f$.**

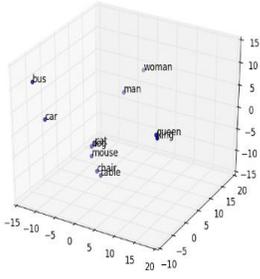

**Figure 2.** Proposed word embedding 3D representation for the words "*dog, mouse, chair, table, car, bus, man, queen, woman, king*".

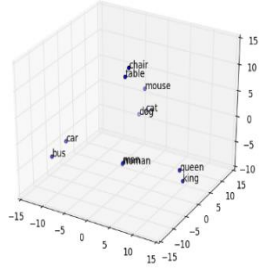

**Figure 3.** Word2Vec word embedding 3D representation for the words "*dog, mouse, chair, table, car, bus, man, queen, woman, king*".

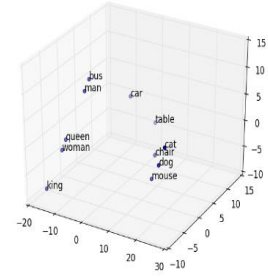

**Figure 4.** Conceptnet word embedding 3D representation for the words "*dog, mouse, chair, table, car, bus, man, queen, woman, king*".

However, the semantic approach only addresses the similarity between words fail to handle the association between words. Association between words is not visible through the context. To address associations as well as similarities that are not visible in context, Conceptnet [21] is used. The set of n related words $w_{c_n}$ to the word w, are extracted from Conceptnet 1. Equation (2) demonstrates the results drawn from the Conceptnet for the word w and generates related words $w_{c_n}$.

$$w_{c_n} = Conceptnet(w) \qquad (2)$$

which consists of $n$ related words extracted from Conceptnet for the given word w since each word w from Conceptnet would hold 20 related words on average, i.e. *n*=20. The extracted words $w_{c_n}$ are converted into vectors $v_{c_n}$ using a pre-trained Word2Vec. The pre-training of Word2Vec pre-training was done on the Wikipedia dataset and create 300 row vector for each word. Each of the extracted words is used to create its own vector (3). The model creates $v_{c_n}$, which is a set of n vectors for each word *w*, and each one of these *n* vectors consists of 300 elements as presented in Word2Vec. The vectors $v_{c_1}$ to $v_{c_n}$ are added together.

$$v_{c_n} = Word2Vec(w_{c_n}) \qquad (3)$$

## 3.2 Embedding Similarity and Association

Adding the extracted vectors $\sum_{i=1}^{n} v_{c_i}$ for the words $w_{c_n}$ creates a generalized 300 row vector representation to the $w$. The combined vector representation is $v_{c_n}$. Extracting words from Conceptnet, would contain similar words and associated words with $w$ [15]. Therefore, adding the vector representations together would generate a generalized vector representation, which holds association and similarity relationships with $w$. Therefore, embedding the associated words and similar words through the use of Conceptnet generates a word embedding which is generalized.

However, a higher weight should be given to semantic word representation directly from the word, w. For this, a Word2Vec representation v is created for the word w directly as well. Equation (4), the v is separately multiplied with the word embedding created through Conceptnet. The vector representation ($v_{cm}$) holds both similarity representation (through the Word2Vec and Conceptnet) and the association representation (through Conceptnet). Therefore, $v_{cm}$ holds deeper relationships that extend the semantic relationships that are extracted and generated from the available context. Therefore, $v_{cm}$ for w would be hold the association relationships and similarity relationships.

$$v_{cm} = \sum_{i=1}^{n} v_{c_i} \cdot v \qquad (4)$$

The $v_{cm}$ is scaled in order to have the word embedding in a reasonable distribution. $v_{cm}$ is further enhanced to generate the final vector representation ($v_f$) using Self Organizing Maps (SOM) [24] that clusters words for the nearest neighbor. The nearest neighbor is identified using the Euclidean distance between the words. The SOM stage is mainly used to optimize the distribution by using SOM. The word embedding could produce a closer relationship within neighbors. The $v_{cm}$ 300-row vector is reshaped to a 2D metrics to pass it to the SOM. $v_{cm}$ is considered a center and the four nearest neighbors based on the Euclidean distance is passed through the SOM. In order to map $v_{cm}$ to the Kohnen's layer, the vector $v_{cm}$ is reshaped into a 2D matrices. The Kohnen's layer would run for 500 iterations, with a learning rate (a) of 0.005. This would produce a new 2D representation for each word. The 2D representation will be transformed into a vector $v_f$ using PCA. Through the use of PCA the dimension reduction is applied to the 2D metric to create the final $v_f$. The $v_f$ is calculated for each word representation $w_{c_n}$ as shown in (5).

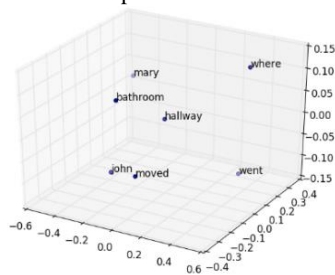

**Figure 5.** The proposed word representation for the context and the question "*Mary moved to the bathroom. John went to the hallway. Where is Mary?*"

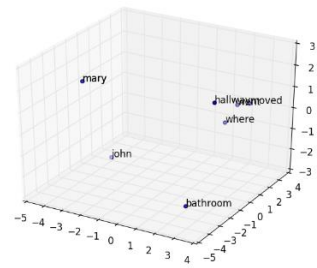

**Figure 6.** Word2Vec representation for the context and question "*Mary moved to the bathroom. John went to the hallway. Where is Mary?*"

---

[1] The related words can be extracted using Conceptnet API. http://api.conceptnet.io/c/en/

$$v_f = \sum_{i=1}^{n}[v_{cm} + a(v_{c_i} - v_{cm})] \quad (5)$$

$v_f$ creates a clear distinction between similar words compared to associated words.

## 4. RESULTS

SimLex-999 is a resource to evaluate models which generate meanings of words and concepts [16]. SimLex-999 captures the similarity between words rather than the relatedness and association. In order to perform achieve a high score, the word embedding should be able to capture similarity independently to the relatedness and association. Therefore, SimLex-999 has shown to be a challenging evaluation model compared to WordSim-353 [25]. The proposed word embedding demonstrates the capability of differentiating similarity from association and relatedness which is reflected in table 1. The context to create the word embedding in order to test on SimLex-999 is created based on Common Crawl data[2]. The proposed word embedding is evaluated using SimLex-999 [16]. SimLex-999 dataset to Spearman's ρ correlation between words is calculated and presented in Table 1. Table 1 compares the proposed model with [1], [2] and [4]. The Spearman's correlation shows that the proposed model is achieving higher Spearman's correlations compared to the semantic word representations. This also shows that the proposed model can differentiate the similarity and association, as the proposed model uses Conceptnet to support generalization and Word2Vec to provide a semantic similarity. We have extended the normal 2D word embedding representation [12] into a 3D word representation using the existing distant measure. The proposed word embedding in a 3D representation is shown in Figure 2 for "dog, mouse, chair, table, car, bus, man, queen, woman, king". The proposed 3D word representations places humans on one plane, vehicles on another plane and objects are placed in another plane with animals but closer to the plane representing humans. Furthermore, the similarities between man to woman is less than queen to king. The proposed word embedding capture more than the gender representation unlike Word2Vec (see Fig. 3). Figure 4 shows the general vector representation created through Conceptnet. This shows a clear representation and is capable of distinguishing association and similarity of the words [16].

3D comparison between the Word2Vec (Fig. 3) and Conceptnet based word embedding (Fig. 4) is shown. The relationship between king to queen and man to woman are similar, but the similarities between them are different as shown in Figure 2 when compared to the word representation using Word2Vec as shown in Figure 3 or using Conceptnet as shown in Figure 4. The vector representation created using only Conceptnet is not context dependent shows a general representation which does not capture similarity between words. This also shows that similarity, association or relatedness cannot be captured by pre-determined relationships which the words hold with each other.

To demonstrate the application of the word representation, for sentences a sample context from the bAbI dataset is used [26]. "*Mary moved to the bathroom. John went to the hallway. Where is Mary?*" is represented in 3D word representation using Word2Vec shown in Figure 5 and the proposed word representation shown in

**Table 1. Performance on SimLexc-999. The proposed word embedding is created using on Wikipedia data. The proposed word embedding and [1] use ≈ 1000m, [2] uses ≈ 990m and [4] uses≈ 852m tokens.**

| Models | Spearman's Correlation |
|---|---|
| Word2Vec [10] | 0.414 |
| Deep Neural Networks with multitask learning | 0.268 |
| Semantic similarity of words using concept networks [24] | 0.76 |
| Human Performance [16] | 0.78 |
| **Proposed word embedding** | **0.886** |

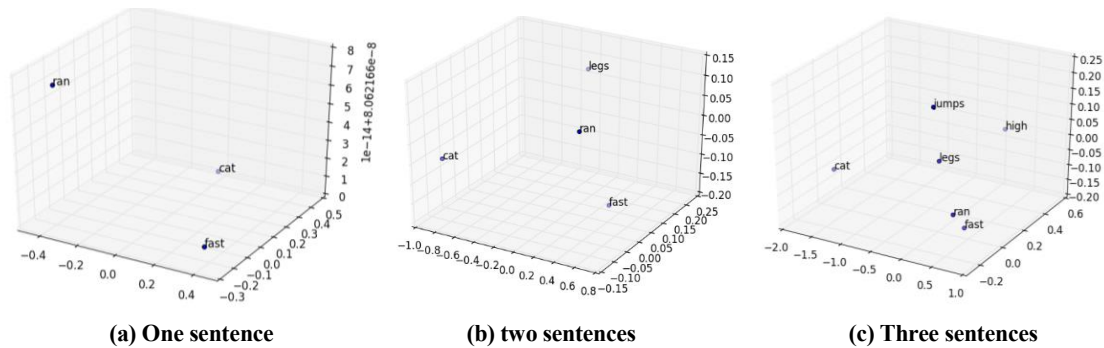

**(a) One sentence**     **(b) two sentences**     **(c) Three sentences**

**Figure 7. The word representation is dynamic. "Cat ran fast. Cat has legs. Cat jumps high". Sentences added one sentence at a time. Capability of achieving similarity and association based on the semantics of the sentences. Word2Vec the semantic representation is captured and Conceptnet generalize vector representation in order to enhance the vector representation. (a) to (c) the vector representation changes with new context. Final representation (c) separation of words in each axis according to the relationship.**

---

[2] https://commoncrawl.org/

Figure 5. Observing the proposed models 3D representations in Figure 5 "*Mary*" and "*bathroom*" are closer in the vector space. Figure 6 demonstrates that Word2Vec places "*John*" is closer to the bathroom compared to "*Mary*". Therefore, the proposed word embedding demonstrates that it supports one factor based question answering tasks. The sentences "*cat ran fast. Cat has legs. Cat jumps high.*" The movement of words in the 3D space when added more sentences shows the dynamics nature of proposed word representation (Figure 7).

## 5. CONCLUSION

This paper presents a word embedding that uses context-based statistics, analogy and deeper related meaning of words to create word representations. The proposed word embedding holds both context-based information via Word2Vec and deep related words via Conceptnet on the words. The word representation uses a higher weight to the context-based information to create the word representation preventing over generalization. The word representation is evaluated using SimLex-999 which achieved a Spearman's correlation of 0.886. Furthermore, the proposed word representations are displayed in 3D representation to show the capability of distinguishing the associated words to similar words based on context. The proposed word embedding is similar to the human performance of word similarity and association by achieving a Spearman's correlation of 0.886 given a human can achieve 0.78 [16].